\documentclass[10pt,twocolumn,letterpaper, pdftex]{article}

\usepackage[pagenumbers]{cvpr} 

\definecolor{cvprblue}{rgb}{0.21,0.49,0.74}
\usepackage[pagebackref,breaklinks,colorlinks,allcolors=cvprblue]{hyperref}
\usepackage{multirow}
\usepackage{colortbl}

\usepackage{algorithm}
\usepackage{algorithmic}

\usepackage{listings}
\usepackage{amsmath}
\usepackage{threeparttable}
\usepackage{wrapfig}
\usepackage{caption}
\usepackage{mathtools}

\def\paperID{1088} 
\def\confName{CVPR}
\def\confYear{2025}

\definecolor{deepgreen}{rgb}{0.0, 0.5, 0.0}

\newcommand{\methodNameNS}{\texttt{OSrCIR}} 
\newcommand{\greycell}{\cellcolor{gray!20}}

\title{Reason-before-Retrieve: One-Stage Reflective Chain-of-Thoughts for Training-Free Zero-Shot Composed Image Retrieval} 

\author{Yuanmin Tang$^{1,2}$\thanks{Work is done during an internship at Microsoft} \quad Xiaoting Qin$^{3}$ \quad Jue Zhang$^{3}$ \quad Jing Yu$^{4}$ \quad Gaopeng Gou$^{1}$ \quad Gang Xiong$^{1}$ \\ 
\quad Qingwei Ling$^{3}$ 
\quad Saravan Rajmohan$^{3}$  \quad Dongmei Zhang$^{3}$   \quad Qi Wu$^{5}$ \\
\textsuperscript{\rm 1}Institute of Information Engineering, Chinese Academy of Sciences \\
\textsuperscript{\rm 2}University of Chinese Academy of Sciences \\
\textsuperscript{\rm 3}Microsoft, \\
\textsuperscript{\rm 4}Minzu University of China, 
\textsuperscript{\rm 5}University of Adelaide \\
    \small{\texttt{\{tangyuanmin,gougaopeng,xionggang\}@iie.ac.cn, jing.emy.yu01@gmail.com, }}\\
    \small{\texttt{\{juezhang,xiaotingqin\}@microsoft.com, qi.wu01@adelaide.edu.au }}
}

\begin{document}
\maketitle
\begin{abstract}
Composed Image Retrieval (CIR) aims to retrieve target images that closely resemble a reference image while integrating user-specified textual modifications, thereby capturing user intent more precisely. Existing training-free zero-shot CIR (ZS-CIR) methods often employ a two-stage process: they first generate a caption for the reference image and then use Large Language Models for reasoning to obtain a target description. However, these methods suffer from missing critical visual details and limited reasoning capabilities, leading to suboptimal retrieval performance. To address these challenges, we propose a novel, training-free one-stage method, One-Stage Reflective Chain-of-Thought Reasoning for ZS-CIR (OSrCIR), which employs Multimodal Large Language Models to retain essential visual information in a single-stage reasoning process, eliminating the information loss seen in two-stage methods. Our Reflective Chain-of-Thought framework further improves interpretative accuracy by aligning manipulation intent with contextual cues from reference images. OSrCIR achieves performance gains of 1.80\% to 6.44\% over existing training-free methods across multiple tasks, setting new state-of-the-art results in ZS-CIR and enhancing its utility in vision-language applications. Our code will be available at \url{https://github.com/Pter61/osrcir2024/}.

\end{abstract}    
\section{Introduction}
\label{sec:intro}

\begin{figure}[t]
    \centering
    \includegraphics[width=1.0\linewidth]{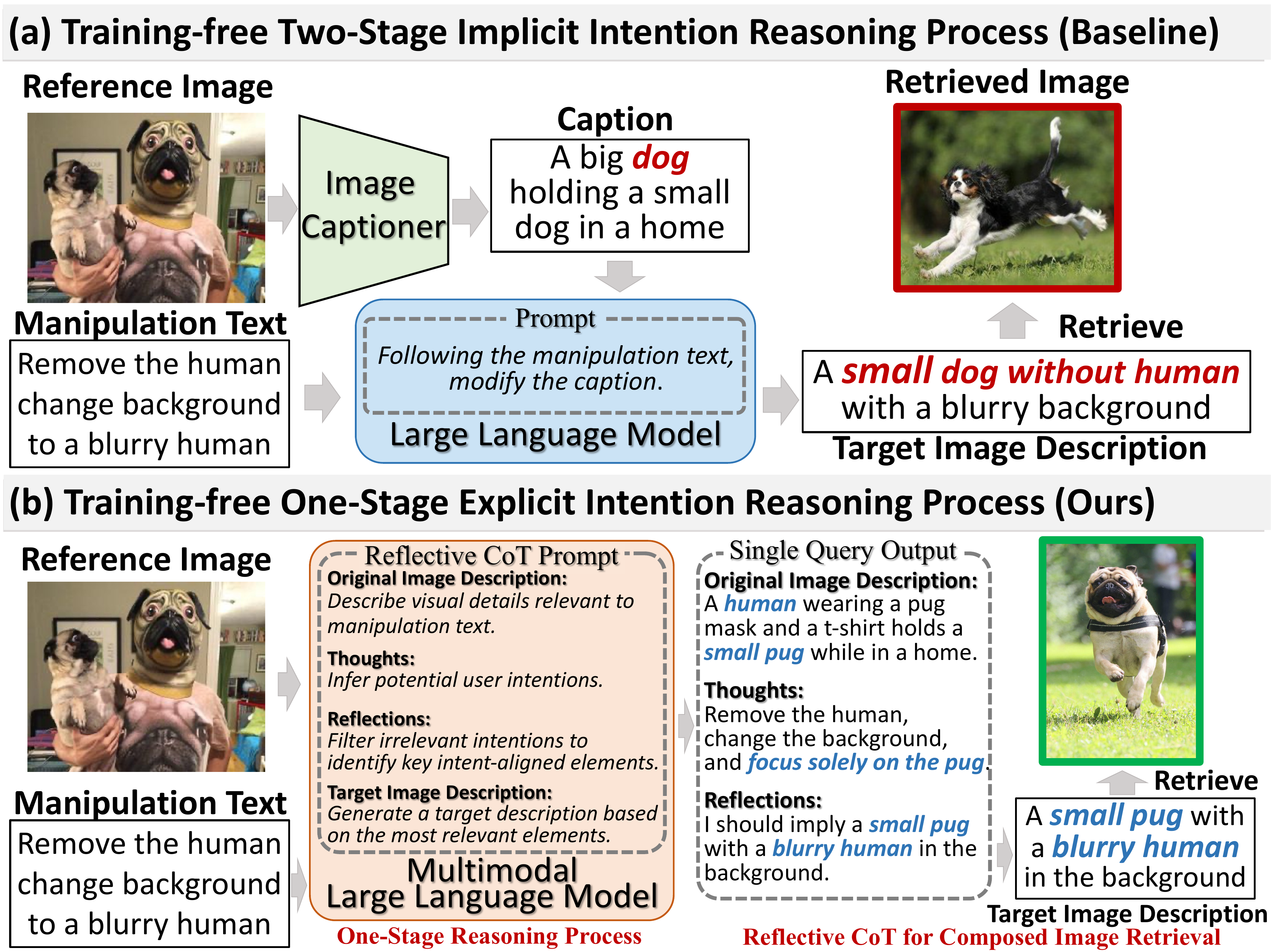}
    \caption{Illustration of our motivation. (a)  Two-stage implicit intention reasoning of the baseline CIReVL method. (b) Our one-stage approach \methodNameNS~with explicit intention reasoning. } 
    \label{fig:motivation}
    \vspace{-10pt}
\end{figure}

Composed Image Retrieval (CIR)~\cite{Vo_2019_CVPR} aims to retrieve a target image that is visually similar to a reference image while incorporating modifications specified by user-provided manipulation text. Unlike traditional content-based image retrieval \cite{datta2008image}, which relies solely on single-modality features, CIR leverages both visual and textual data to capture user intent more accurately, as shown in Figure~\ref{fig:motivation}. This dual-modality approach allows users to specify desired changes to reference images, improving search precision and enabling a clearer articulation of user intent. Consequently, CIR has garnered increasing interest in internet search and e-commerce~\cite{Chen_2020_CVPR, Saito_2023_CVPR}, where it facilitates tasks such as scene image search with object manipulation or product recommendations with attribute modification.

CIR faces two fundamental challenges: (1) user intent spans both visual and textual modalities, necessitating a common semantic space for effective cross-modal reasoning, and (2) understanding user intent demands deep reasoning, as it is often implicitly conveyed, particularly through reference images. While supervised methods have been proposed to tackle these issues~\cite{Liu_2021_ICCV, Baldrati_2022_CVPR}, they rely on extensive annotated triplets (\textit{i.e.,} reference image, manipulation text, target image) CIR datasets to train task-specific models, which is labor-intensive and limits generalizability.

Zero-Shot Composed Image Retrieval (ZS-CIR) has emerged as a solution to these limitations~\cite{Saito_2023_CVPR, baldrati2023zero, tang2023contexti2w}, utilizing the pre-trained large-scale Vision-Language Models (VLMs), \textit{i.e.,} CLIP~\cite{radford2021learning}, to reframe ZS-CIR as a text-based image retrieval task. It encodes reference image content into language and combines them with manipulation text to obtain query captions for target retrieval within CLIP's shared semantic space. Query generation methods in ZS-CIR can be implicit or explicit. Implicit methods, like textual inversion~\cite{Saito_2023_CVPR, baldrati2023zero, tang2023contexti2w}, are often training-dependent, using large image-caption datasets to train a mapping network that converts images into text tokens.  A static template then combines these tokens with textual modifications to create query captions. However, even with large-scale VLMs, these implicit ZS-CIR methods are limited by CLIP capacity for human intention reasoning, which restricts the accurate interpretation of manipulation intent.

Alternatively, recent research~\cite{sun2023training, karthik2024visionbylanguage} explores training-free ZS-CIR methods that utilize Large Language Models (LLMs) for explicit query inference. As illustrated in Figure \ref{fig:motivation}(a), current explicit training-free methods follow a two-stage process: an image captioner (\textit{e.g.,} BLIP-2 \cite{li2023blip2}) first encodes the reference image into text, followed by LLM-based reasoning to derive a target image description for retrieval. Despite this progress, current two-stage LLM-based methods for ZS-CIR still face two limitations:

(1) \textbf{Missing Visual Information.} The initial captioning process is not informed by manipulation text, so critical visual details needed for query composition are often missing. For instance, in Figure~\ref{fig:motivation}, without explicit emphasis on the term ``human'' in manipulation text, the caption fails to include the term ``human holds pug''. 
Thus, even with a large-scale retrieval model, this problem remains unresolved.

(2) \textbf{Limited Exploitation of LLM Reasoning Capabilities.} Although LLMs offer strong reasoning capabilities, current methods often rely on simple reasoning prompts like \textit{following $<$Manipulation Text$>$, modify $<$Caption$>$} \cite{karthik2024visionbylanguage}, which restricts LLMs' full reasoning potential and may lead to suboptimal inferences. As seen in Figure~\ref{fig:motivation}, the true user intent of ``a blurry human in the background'' is misinterpreted as ``without human with a blurry background''.

To address these limitations, we propose a novel training-free \textit{\textbf{O}ne-\textbf{S}tage \textbf{r}eflective chain-of-thought reasoning for zero-shot \textbf{C}omposed \textbf{I}mage \textbf{R}etrieval} (\textbf{OSrCIR}). As shown in Figure \ref{fig:motivation}(b), in this one-stage reasoning process, we leverage Multimodal Large Language Models (MLLMs) that handle visual and textual inputs simultaneously, thereby avoiding the intrinsic information loss seen in two-stage methods. Our Reflective Chain-of-Thought (CoT) framework further enhances reasoning by interpreting nuanced manipulation intents from both the manipulation text and contextual cues in the reference images, allowing the model to more accurately locate and apply relevant visual details. This approach is inspired by human cognitive processes, particularly iterative refinement and reasoning, enhancing both model performance and interpretability. 

The main contributions are summarized as follows: (1) We propose a one-stage reasoning method based on MLLMs, which fully retains the visual information of the reference image. This approach helps unleash the model's reasoning ability in CIR, thereby improving the accuracy and efficiency of training-free ZS-CIR. (2) We designed a Reflective CoT reasoning approach to address the current model's insufficient understanding of manipulation intention. This approach interprets visual intent based on visual information and accurately identifies relevant visual elements during reasoning, significantly enhancing model performance and interpretability. (3) Our model improves from 1.80\% to 6.44\% across four tasks on ViT-L/14 while maintaining inference efficiency, setting new state-of-the-art results in ZS-CIR, further impacting a broader range of vision and language applications.

\section{Related works}
\label{sec:related_works}

\textbf{Composed Image Retrieval.} Composed Image Retrieval (CIR) involves combining image and text features for retrieval \cite{Vo_2019_CVPR}, using late fusion to integrate visual and textual features while requiring extensive annotated triplets CIR datasets \cite{Baldrati_2022_CVPR, Liu_2021_ICCV, zhang2024magiclens}. Zero-shot CIR models \cite{Saito_2023_CVPR, baldrati2023zero, tang2023contexti2w, gu2024lincir, karthik2024visionbylanguage,tang2024denoise,Suo_2024_CVPR,du2024image2sentence,FTI4CIR,jang2024spherical,tang2024manipulation} eliminate the need for large-scale CIR datasets enabling CIR without extensive labeled data. Textual inversion ZS-CIR methods \cite{Baldrati_2022_CVPR,baldrati2023zero,gu2024lincir} leverages image-text pairs during training, using pre-trained CLIP language encoder for reasoning. However, these methods often struggle to interpret implicit human intent embedded in manipulation text. Training-free ZS-CIR approaches \cite{karthik2024visionbylanguage, sun2023training, yang2024ldre}, such as CIReVL \cite{karthik2024visionbylanguage}, leverage LLM to infer manipulation intent. However, their two-stage process, where image captioning is conducted independently of the manipulation text, often results in inaccuracies, as critical visual details and implicit intent are missed. To address these challenges, we propose a one-stage approach that directly reasons about user intent using complete image content. Unlike diffusion-based \cite{gu2023compodiff} or ensemble-based methods like LDRE \cite{yang2024ldre}, which introduce substantial computational overhead, our model achieves greater efficiency and faster inference times.

\begin{figure*}
    \centering
    \includegraphics[width=1.0\linewidth]{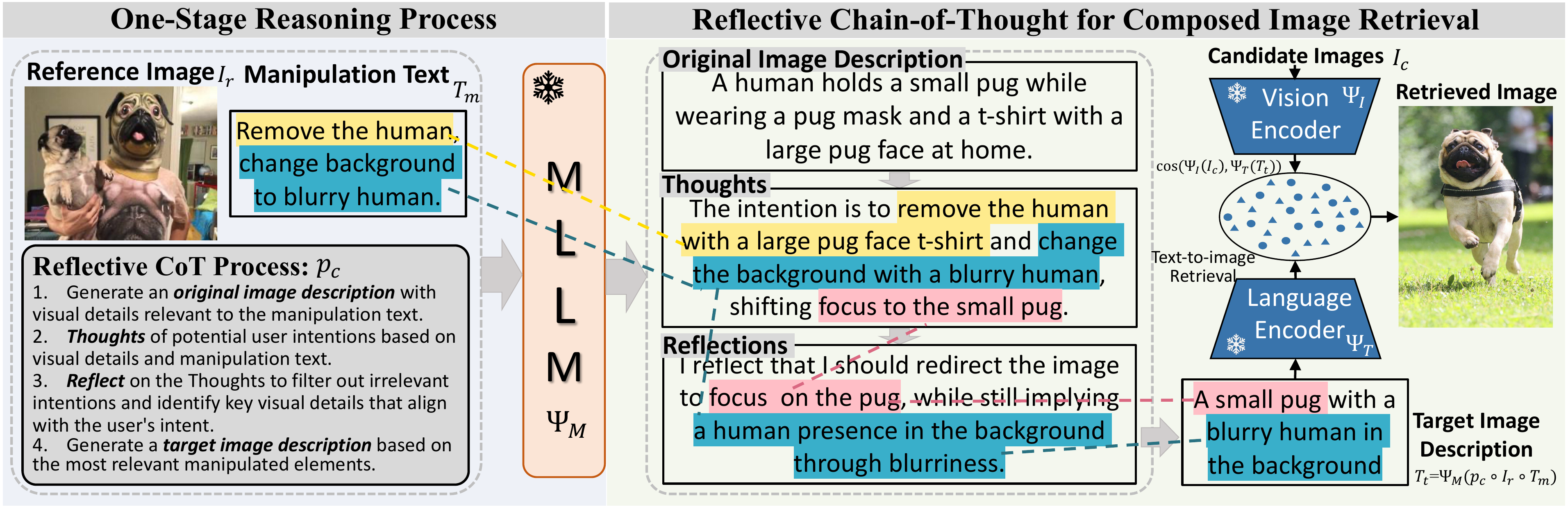}   
    \caption{An overview of our model.  An MLLM processes the reference image and the manipulation text to generate a description of the desired target image by reflective CoT. To obtain the desired image, we use a vision-language model and perform text-to-image retrieval. Different colors denote the reasoning outcomes of each intention.  }
    \label{fig:model-architecture}
    \vspace{-10pt}
\end{figure*}

\noindent\textbf{Vision and Language Pre-training Models.} Vision and Language Pre-training (VLP) models, such as CLIP \cite{radford2021learning}, leverage large-scale image-text pairs to align visual and textual data implicitly. Recent advancements in VLP \cite{Zhou_2022_CVPR, song2022clip} have employed static models that merge encoded image and text features, enabling a variety of zero-shot tasks \cite{pmlr-v162-li22n, NEURIPS2022_960a172b, li2023blip2, song2022clip, hummel2024egocvr, Shi_2023_ICCV,Shi_2024_ECCV}. More recent work has focused on integrating vision and language processing within the architecture of large pre-trained language models, leading to the development of state-of-the-art Multimodal Large Language Models (MLLMs) such as LLaVA~\cite{llava2023neurips} and GPT-4~\cite{GPT-4V, GPT-4o}, which offer enhanced multimodal capabilities. Additionally, methods like PVIT~\cite{Chi23imageretrieval}, GRACE~\cite{Li24crossmodal}, LightningDOT~\cite{lightningdot}, and ComCLIP~\cite{comclip} have further enhanced the cross-modal retrieval capabilities of multimodal models, pushing the boundaries of image-text matching and retrieval tasks. Our work demonstrates that an MLLM alone, when combined with vision-language retrieval models, can suffice for effective CIR without additional training.

\noindent\textbf{Reasoning Capability of LLMs and MLLMs.} LLMs demonstrate strong reasoning abilities, largely enabled by in-context learning (ICL) \cite{brown2020language}, where prompted examples and contextual cues improve model performance. Chain-of-Thought prompting \cite{wei2022chain} further enhances reasoning by guiding LLMs to generate intermediate reasoning steps in complex reasoning tasks. Studies show that LLMs benefit from both crafted demonstrations \cite{wei2022chain} and zero-shot prompting \cite{kojima2022large}. Furthermore, self-reflection techniques \cite{reflection23neurips} have proven effective in enhancing reasoning, as they allow models to assess and refine their outputs iteratively. However, MLLMs face challenges in reasoning due to the gap between visual and textual data. To address this gap, recent research has developed advanced training \cite{NEURIPS2022_960a172b, peng2023kosmos, luo2020multi, zhu2022seqtr} and prompting methods \cite{hong2023cogagent, zhang2024makes, zhang2023prompt, Mitra2024composalCoT, zheng2023ddcot}. Several studies \cite{mmcot2023journal, MMLatentCoT23, zheng2023ddcot, cantor24cot, zhang24cocot, Wang2024TSciQCoT, Mitra2024composalCoT} have adapted CoT for multimodal reasoning tasks, such as visual question answering \cite{antol2015vqa}, showing that CoT can significantly enhance visual reasoning in MLLMs. Building on these advancements, our work is the first to apply CoT to ZS-CIR, extending CoT's impact to a new multimodal domain.
\section{Methodology}
\label{sec:method}

Given a reference image \( I_{r} \) and a manipulation text \( T_{m} \) describing the user's intention of hypothetical semantic changes on the reference image, Zero-Shot Composed Image Retrieval (ZS-CIR) retrieves images from an image database  \( \mathcal{D}\) that are visually similar to \( I_{r} \) while incorporating the modifications specified in \( T_{m} \). Figure \ref{fig:model-architecture} illustrates our model. We introduce a new approach to explicitly reasoning a target image description \( T_{t} \) as the composed query based on a Multimodal Large Language Model (MLLM) \( \Psi_M \), which contains pre-trained knowledge to understand the user's intention embedded in manipulation text. 
To ensure that \( \Psi_M \) reasons \( T_{t} \) in a human-understandable manner, we introduce a Reflective Chain-of-Thought prompt $p_c$.
The obtained target image description $T_m$ is then used for image retrieval via CLIP, with the associated pre-trained text encoder \( \Psi_T \) embedding both the target image description \( T_t \) and candidate images \( I_c \) into a shared, searchable space. The matching score is computed using cosine similarity \( \texttt{cos}(\Psi_I(I_{c}), \Psi_T(T_t)) \).

\subsection{One-Stage Reasoning Process}
The conventional two-stage structure of training-free ZS-CIR restricts the ability of image captioners to capture essential visual details, thereby constraining the reasoning capacity of LLMs. To overcome this limitation, we propose a streamlined one-stage approach that eliminates the need for a separate image captioning stage, which does not include user provided manipulation intent. As shown in Figure~\ref{fig:model-architecture} (left), we aim to leverage \( \Psi_M \)'s inherent multimodal understanding to capture the reference image's details directly. This enables reasoning a target image description \( T_{t} \), modeling the user's intention of hypothetical manipulation of \( T_{m} \) on the reference image \( I_{r} \) as a transformation in the resulting target description \( T_{t} \) without additional training.
Formally, given an MLLM \( \Psi_M \), we generate a target image description  \( T_{t} \) contains the user's manipulation intent \( T_{m} \) on the reference image \( I_{r} \) as follows:
\begin{equation}
\begin{aligned}
T_{t} = \Psi_M(p_c \circ I_r \circ T_{m}),
\end{aligned}
\label{f:tgt}
\end{equation}
where the LLM is queried with a concatenated prompt composed of the base CoT prompt \( p_{c} \) (see Section \ref{secc:Reflective_CoT} for details), the reference image \( I_r \) (prepended with ``\texttt{Original Image Context}''), and \( T_{m} \), the manipulation intent text (prepended with ``\texttt{Manipulation text}''). This prompt format is largely task-agnostic, enabling its application across a variety of CIR tasks.

\subsection{Reflective Chain-of-Thought for ZS-CIR} 
\label{secc:Reflective_CoT}

Each image-intention input pair comprises a reference image and manipulation text that implicitly conveys the user's intention to modify the reference image. To generate the target image description $T_t$, the adopted MLLM needs to understand this manipulation intention accurately. 
Existing methods rely on simple prompts (\textit{e.g.,} \texttt{Following \( T_{m} \), modify reference image caption}) to extract these intentions, but this approach is insufficient for accurately inferring user's implicit intention embedded in \( T_{m} \) (see Section \ref{sec::anaysis}). To address this limitation, we introduce a Reflective CoT prompt \( p_c \), which guides the MLLM to progressively reason about user intent across both the reference image and manipulation text, ensuring accurate ZS-CIR.

Specifically, as shown in Figure~\ref{fig:model-architecture} (right), the Reflective CoT prompt instructs the following progressive reasoning steps: First, the \textit{Original Image Description} step highlights visual details relevant to the user's intention in the reference image. The \textit{Thoughts} step then captures the user's intention and reasoning for potentially manipulated visual elements. In the \textit{Reflections} step, these elements are further evaluated to identify those mostly aligned with the user's intent. Finally, the \textit{Target Image Description} step generates a refined description based on the most intention-relevant visual modifications for target retrieval. Notably, all steps are included in a \textbf{single} prompt for MLLM, ensuring both efficiency and interpretability. We illustrate each reasoning step below using the example in Figure~\ref{fig:model-architecture}, while providing the complete prompt template in Appendix \ref{sec::tem}.

\noindent \textbf{Original Image Description.} During this step, the MLLM is asked to \textit{capture all visible objects, attributes, and elements relevant to the manipulation text}, and to \textit{reflect on the content and context of the image} to ensure retention of fine-grained details. In Figure~\ref{fig:model-architecture}, the intention-irrelevant visual details (\textit{e.g.,} a table, lights, or photos) are excluded in the caption while relevant elements (\textit{e.g.,} human holding a small pug) are preserved to align with the manipulation text.

\noindent \textbf{Thoughts.} Given the intention-relevant visual details and manipulation text, the MLLM then seeks to capture the user's intention (\textit{e.g.,} ``Remove the human, change the background''). We first prompt the MLLM to \textit{explain its understanding of the manipulation intent}. Since the user’s intentions are often implicit, requiring reference image context for interpretation (\textit{e.g.,} ``Removing the human to focus on the pug''), we further ask the MLLM to \textit{discuss how the manipulation intent influences the choice of focused elements in the original image}.

\noindent \textbf{Reflections.} Given the manipulation intent and reference image, the MLLM needs to filter out incorrect intentions (\textit{e.g.,} removing the human) and identify the most relevant manipulated elements (\textit{e.g.,} the small pug, a blurry human background). We ask the MLLM to \textit{highlight key decisions made to preserve the coherence and context of the original image while fulfilling the manipulation intent} and to \textit{offer a logical connection between the original content and the final description.} This step also alleviates hallucination issues present in the Thoughts step (See Figure \ref{fig:CoT}).

\noindent \textbf{Target Image Description.} Given the filtered manipulated elements, the MLLM finally generates a target description based on the manipulated elements mostly relevant to user intent. We simply ask the MLLM to \textit{generate a target image description that only contains the target content}.

\begin{table*}[t!]
    \centering
    \resizebox{0.75\linewidth}{!}{
    \begin{tabular}{l|l|cccc||ccc|ccc}
    \toprule
    \multicolumn{2}{c}{\textbf{CIRCO + CIRR $\rightarrow$}} & \multicolumn{4}{|c||}{\textbf{CIRCO}}& \multicolumn{6}{|c}{\textbf{CIRR}}\\
    \midrule
    \multicolumn{2}{c}{Metric} & \multicolumn{4}{|c||}{mAP@k}& \multicolumn{3}{|c|}{Recall@k}& \multicolumn{3}{c}{Recall$_{\text{Subset}}$@k}\\
     Arch & Method & k=5 & k=10 & k=25 & k=50 & k=1 & k=5 & k=10 & k=1 & k=2 & k=3 \\
     \midrule
     \multirow{4}{*}{ViT-B/32} & SEARLE  & 9.35 & 9.94 & 11.13 & 11.84& 24.00 & 53.42 & 66.82 & 54.89 & 76.60 & 88.19 \\
       & \greycell CIReVL & \greycell14.94 & \greycell15.42 & \greycell17.00 & \greycell17.82& \greycell23.94 & \greycell52.51 & \greycell66.00 & \greycell60.17 & \greycell80.05 & \greycell90.19 \\
      & \greycell CIReVL$^{*}$ & \greycell \underline{16.02} & \greycell \underline{16.69} & \greycell \underline{17.77} & \greycell \underline{18.89} & \greycell \underline{24.25} & \greycell \underline{52.83} & \greycell\underline{66.32} & \greycell\underline{60.43} & \greycell\underline{80.35} & \greycell \underline{90.51} \\
     & \greycell \textbf{\methodNameNS} & \greycell \textbf{18.04} & \greycell \textbf{19.17} & \greycell \textbf{20.94} & \greycell \textbf{21.85} & \greycell \textbf{25.42} & \greycell \textbf{54.54} & \greycell \textbf{68.19} & \greycell \textbf{62.31} & \greycell \textbf{80.86} & \greycell \textbf{91.13} \\
     
    \midrule
     \multirow{7}{*}{ViT-L/14} & Pic2Word  & 8.72 & 9.51 & 10.64 & 11.29 & 23.90 & 51.70 & 65.30 & - & - & - \\
      & SEARLE  & 11.68 & 12.73 & 14.33 & 15.12 & 24.24 & 52.48 & 66.29 & 53.76 & 75.01 & 88.19\\
      & LinCIR  & 12.59  & 13.58  & 15.00  & 15.85 & 25.04  & 53.25  & 66.68 & 57.11  & 77.37  & 88.89 \\
      & Context-I2W  & 13.04  & 14.62  & 16.14  & 17.16 & \underline{25.60}  & \underline{55.10}  & \underline{68.50} & -  & -  & - \\
      & \greycell CIReVL & \greycell 18.57 & \greycell 19.01 & \greycell 20.89 & \greycell 21.80 & \greycell 24.55 & \greycell 52.31 & \greycell 64.92 & \greycell 59.54 & \greycell 79.88 & \greycell 89.69\\
        & \greycell CIReVL$^{*}$ & \greycell \underline{18.92} & \greycell  \underline{19.32} & \greycell \underline{21.15} & \greycell \underline{22.14} & \greycell 24.83 & \greycell 52.68 & \greycell 65.28 & \greycell \underline{59.82} & \greycell \underline{80.15} & \greycell \underline{89.98}\\
      & \greycell \textbf{\methodNameNS} & \greycell \textbf{23.87} & \greycell \textbf{25.33} & \greycell \textbf{27.84} & \greycell \textbf{28.97} & \greycell \textbf{29.45} & \greycell \textbf{57.68} & \greycell \textbf{69.86} & \greycell \textbf{62.12} & \greycell \textbf{81.92} & \greycell \textbf{91.10} \\
     \midrule
      \multirow{4}{*}{ViT-G/14} & LinCIR & 19.71 & 21.01 & 23.13 & 24.18 & \underline{35.25} & \underline{64.72} & \underline{76.05} & 63.35  & 82.22  & 91.98\\
      & \greycell CIReVL & \greycell 26.77 & \greycell 27.59 & \greycell 29.96 & \greycell 31.03& \greycell 34.65 & \greycell 64.29 & \greycell 75.06 & \greycell 67.95 & \greycell 84.87 & \greycell 93.21\\
      & \greycell CIReVL$^{*}$ & \greycell \underline{27.12} & \greycell \underline{28.01} & \greycell \underline{30.35} & \greycell \underline{31.39} & \greycell 34.98 & \greycell 64.68 & \greycell 75.41 & \greycell \underline{68.37} & \greycell \underline{85.23} & \greycell \underline{93.24} \\
   & \greycell \textbf{\methodNameNS} & \greycell \textbf{30.47} & \greycell \textbf{31.14} & \greycell \textbf{35.03} & \greycell \textbf{36.59} & \greycell \textbf{37.26} & \greycell \textbf{67.25} & \greycell \textbf{77.33} & \greycell \textbf{69.22}  & \greycell \textbf{85.28}  & \greycell \textbf{93.55}\\
    \bottomrule
    \end{tabular}}   
    \vspace{-3pt}
    \caption{\textbf{Comparison on CIRCO and CIRR Test Data.} On CIRCO, \methodNameNS~significantly outperforms even adaptive methods across retrieval models, while it achieves competitive results on CIRR despite the noise in the benchmark. Grey lines represent the training-free ZS-CIR methods. CIReVL$^*$ uses the GPT4o \cite{achiam2023gpt} in two-stage. \textbf{Bold} and `$\underline{\phantom{x}}$' denote the best and second-best result, respectively.} 
    \label{tab:circo}
    \vspace{-13pt}
\end{table*}

\noindent \textbf{Vision-by-Language In-Context Learning.} Simply providing guidelines for the Reflective CoT process is insufficient for MLLMs to understand the CoT process required at each step. To address this, we leverage in-context learning, a technique widely used in LLM and MLLM CoT methods \cite{wei2022chain, mitra2023compositional, zheng2023ddcot}. To ensure a zero-shot setting in ZS-CIR, we propose a vision-by-language in-context learning (ICL) approach. This method provides a few expected MLLM outputs in text form as examples, without requiring a reference image, to guide the MLLM through the reasoning process at each step. Refer to in the Appendix \ref{sec::ICT_details} for more details.

\noindent \textbf{Composed Image Retrieval.} Given the target image description \( T_t \), our model encodes the image-search database \( \mathcal{D} \) alongside \( T_t \) using a frozen pre-trained CLIP. The retrieved target image \( I_t \) is determined as follows:
\begin{equation}
\begin{aligned}
\label{eq:retrieval}
I_t = \underset{I_r\in\mathcal{D}}{\mathtt{argmax}}\; \frac{{\Psi_I(I_r)}^\intercal \Psi_T(T_t)}{||\Psi_I(I_r)|| ~ ||\Psi_T(T_t)||},
\end{aligned}
\end{equation}
where the selected target image \( I_t \) is the one most similar to the generated target image description. The retrieval process is modular, performed only after combining the reference image and manipulation text, allowing flexibility to substitute different retrieval systems based on practical needs and the desired trade-off between efficiency and effectiveness. Our approach enables a human-understandable ZS-CIR pipeline, where reasoning is fully expressed in the language domain, and the retrieval process is clearly separated, requiring no additional training or mapping modules.

\section{Experiments}
\label{sec:exp}

\begin{figure}[t]
    \centering
    \centering
    \includegraphics[width=1.0\linewidth]{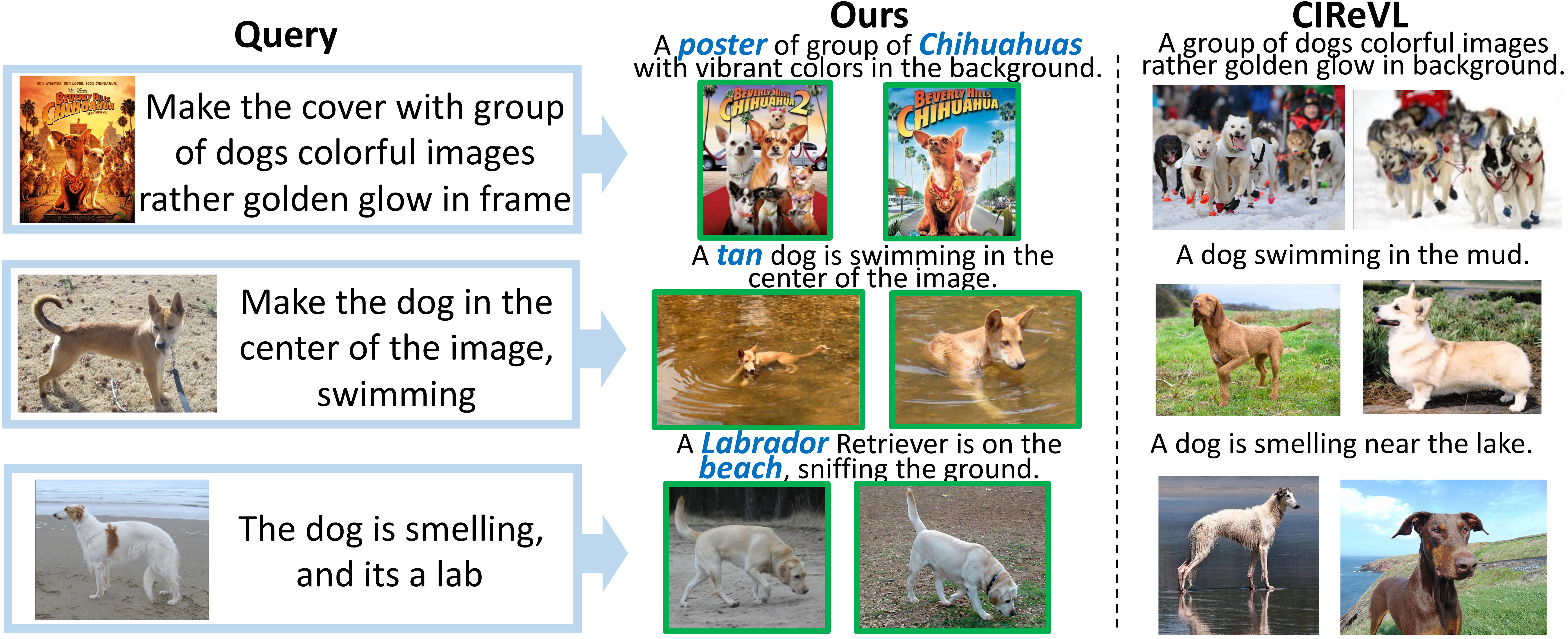}
    \caption{Results on the object manipulation on the CIRR.}
    \label{fig:cirr}
    \vspace{-13pt}
\end{figure}

\noindent\textbf{Datasets and Baselines.} We utilize four commonly used datasets in CIR: CIRR~\cite{Liu_2021_ICCV}, CIRCO~\cite{baldrati2023zero}, FashionIQ~\cite{Wu_2021_CVPR}, and GeneCIS~\cite{vaze2023genecis}. CIRR is the first natural image dataset for CIR, although it can include false negatives~\cite{baldrati2023zero}, where several images could be potential ground truths but are not labeled as such. The CIRCO dataset addresses this by providing multiple annotated ground truths to reduce false negatives. GeneCIS, built from MS-COCO~\cite{coco} and Visual Attributes in the Wild~\cite{vaw}, offers four task variations, enabling retrieval or modification tasks around specific objects or attributes. FashionIQ focuses specifically on fashion-related retrieval. 
These datasets cover distinct CIR tasks: CIRCO and CIRR for object manipulation (using reference images to guide object or background manipulation), GeneCIS for object and attribute composition (with various object and attribute labels used to combine with cropped query images for retrieval), and FashionIQ for attribute manipulation (offering descriptive sentences to modify image attributes).
Following the original benchmarks, we use Recall@k (R@k) as the evaluation metric for CIRR, GeneCIS, and FashionIQ, and mean average precision (mAP@k) for CIRCO to account for multiple positives. We also evaluate CIRR in a subset setting, where Recall$_\text{Subset}$@k measures retrieval performance within a limited selection of images relevant to the query in the database.

\begin{table*}[t!]
    \centering  
    \resizebox{0.85\linewidth}{!}{
    \begin{tabular}{l|l|cccccccccccc|c}
    \toprule
    \multicolumn{2}{c|}{\textbf{GeneCIS $\rightarrow$}} & \multicolumn{3}{c}{Focus Attribute} & \multicolumn{3}{c}{Change Attribute} & \multicolumn{3}{c}{Focus Object} & \multicolumn{3}{c|}{Change Object} & \textbf{Average}\\
    \midrule
     Backbone & Method & R@1 & R@2 & R@3 & R@1 & R@2 & R@3 & R@1 & R@2 & R@3 & R@1 & R@2 & R@3 & R@1 \\
     \midrule
     \multirow{4}{*}{ViT-B/32}& SEARLE & \underline{18.9} & \underline{30.6} & \underline{41.2} & 13.0 & 23.8 &33.7 & 12.2 &23.0 & 33.3 & 13.6 &	23.8 &	33.3 &	14.4 \\
      & \greycell CIReVL & \greycell 17.9 & \greycell 29.4 & \greycell 40.4 & \greycell 14.8 & \greycell 25.8 & \greycell 35.8 & \greycell 14.6 & \greycell 24.3 & \greycell 33.3 & \greycell 16.1 & \greycell 27.8 & \greycell 37.6 & \greycell 15.9 \\
    & \greycell CIReVL$^{*}$ & \greycell  18.2 & \greycell 29.7 & \greycell 40.7 & \greycell \underline{15.1} & \greycell \underline{26.1} & \greycell \underline{36.1} & \greycell \underline{14.9} & \greycell \underline{24.5} & \greycell \underline{33.6} & \greycell \underline{16.4} & \greycell \underline{28.1} & \greycell \underline{37.9} & \greycell \underline{16.2} \\
      & \greycell \textbf{\methodNameNS} & \greycell \textbf{19.4} & \greycell \textbf{32.7} & \greycell \textbf{42.8} & \greycell \textbf{16.4} & \greycell \textbf{27.7} & \greycell \textbf{38.1} & \greycell \textbf{15.7} & \greycell \textbf{25.7} & \greycell \textbf{35.8} & \greycell \textbf{18.2} & \greycell \textbf{30.1} & \greycell \textbf{39.4} & \greycell \textbf{17.4} \\
     \midrule
      \multirow{6}{*}{ViT-L/14}& SEARLE & 17.1 &	29.6 & 40.7 & 16.3 &	25.2 & 34.2 &	12.0 & 22.2 & 30.9 & 12.0 & 24.1 & 33.9 & 14.4\\
      & LinCIR & 16.9  & 30.0  & 41.5 & 16.2  & 28.0  & 36.8 & 8.3  & 17.4  & 26.2 & 7.4  & 15.7  & 25.0 & 12.2 \\
      & Context-I2W & 17.2  & 30.5  & 41.7 & \underline{16.4}  & \underline{28.3}  & \underline{37.1} & 8.7  & 17.9  & 26.9 & 7.7  & 16.0  & 25.4 & 12.7 \\
      & \greycell CIReVL & \greycell 19.5 & \greycell 31.8 & \greycell 42.0 & \greycell 14.4 & \greycell 26.0 & \greycell 35.2 & \greycell 12.3 & \greycell 21.8 & \greycell 30.5 & \greycell 17.2 & \greycell 28.9 & \greycell 37.6 & \greycell 15.9\\
      & \greycell CIReVL$^{*}$ & \greycell \underline{19.7} & \greycell \underline{32.1} & \greycell \underline{42.3} & \greycell 14.8 & \greycell 26.2 & \greycell 35.4 & \greycell \underline{12.5} & \greycell \underline{22.1} & \greycell \underline{30.7} & \greycell \underline{17.3} & \greycell \underline{29.1} & \greycell \underline{37.9} & \greycell \underline{16.1} \\
      & \greycell \textbf{\methodNameNS} & \greycell \textbf{20.9} & \greycell \textbf{33.1} & \greycell \textbf{44.5} & \greycell \textbf{17.2} & \greycell \textbf{28.5} & \greycell \textbf{37.9} & \greycell \textbf{15.0} & \greycell \textbf{23.6} & \greycell \textbf{34.2} & \greycell \textbf{18.4} & \greycell \textbf{30.6} & \greycell \textbf{38.3} & \greycell \textbf{17.9} \\
     \midrule
     \multirow{4}{*}{ViT-G/14} & LinCIR & 19.1  & 33.0  & 42.3 & \underline{17.6} & \underline{30.2} & \underline{38.1} & 10.1  & 19.1  & 28.1 & 7.9  & 16.3  & 25.7 & 13.7 \\
     & \greycell CIReVL & \greycell 20.5 & \greycell 34.0 & \greycell 44.5 & \greycell 16.1 & \greycell 28.6 & \greycell 39.4 & \greycell 14.7 & \greycell 25.2 & \greycell 33.0 & \greycell 18.1 & \greycell 31.2 & \greycell 41.0 & \greycell 17.4\\
     & \greycell CIReVL$^{*}$ & \greycell \underline{20.9} & \greycell \underline{34.4} & \greycell \underline{44.9} & \greycell 16.5 & \greycell 29.0 & \greycell 39.8 & \greycell \underline{15.1} & \greycell \underline{25.6} & \greycell \underline{33.4} & \greycell \underline{18.5}  & \greycell \underline{31.6} & \greycell \underline{41.4} & \greycell \underline{17.8}\\
     & \greycell \textbf{\methodNameNS} & \greycell \textbf{22.7} & \greycell \textbf{36.4} & \greycell \textbf{47.0} & \greycell \textbf{17.9} & \greycell \textbf{30.8} & \greycell \textbf{42.0} & \greycell \textbf{16.9} & \greycell \textbf{28.4} & \greycell \textbf{36.7} & \greycell \textbf{21.0} & \greycell \textbf{33.4} & \greycell \textbf{44.2} & \greycell \textbf{19.6}\\
    \bottomrule
    \end{tabular}}    
    \vspace{-5pt}
    \caption{\textbf{Comparison on GeneCIS Test Data.} \methodNameNS~is able to significantly outperform adaptive methods across all GeneCIS sub-benchmarks, with its inherent modularity allowing for further simple scaling to achieve additional large gains. Grey lines represent the training-free ZS-CIR methods. CIReVL$^*$ uses the GPT4o in two-stage. \textbf{Bold} and `$\underline{\phantom{x}}$' denotes the best and second-best result, respectively.}
    \label{tab:genecis}
    \vspace{-10pt}
\end{table*}

We compare \methodNameNS~with several commonly benchmarked ZS-CIR methods, categorized as textual inversion or training-free approaches. The textual inversion methods are training-dependent and include: 1) \textbf{Pic2Word} \cite{Saito_2023_CVPR}: maps the visual features of a reference image into a pseudo-word token. 2) \textbf{SEARLE} \cite{baldrati2023zero}: combines the pseudo-word token with the GPT-generated caption~\cite{brown2020language} and applies distillation for efficiency. 3) \textbf{Context-I2W} \cite{tang2023contexti2w}: selectively maps text description-relevant visual information from the reference image. 4) \textbf{LinCIR} \cite{gu2024lincir}: masks subjects in captions to enhance training efficiency. 

The training-free baseline methods are as follows: 1) \textbf{CIReVL} \cite{karthik2024visionbylanguage}, a two-stage approach where a pre-trained image captioner generates a reference image caption, followed by an LLM composing a target image description based on manipulation text; and 2) \textbf{CIReVL$^{*}$}, following CIReVL's two-stage process but employing the same MLLM used in \methodNameNS~for both reference image captioning and target image description generation.
To ensure a fair comparison, we present results without using LLM-based ensemble methods like LDRE~\cite{yang2024ldre} or diffusion-based models like CompoDiff \cite{gu2023compodiff}, as these approaches add substantial computational overhead in inference or training. We evaluate our method across three backbones (ViT-B/32, ViT-L/14, and ViT-G/14) but focus primarily on ViT-L/14 for baseline comparisons. This choice is driven by its balance of inference efficiency and retrieval quality, which is widely reported by other baselines and is more practical for real-world applications.

\noindent\textbf{Implementation Details.} 
The default MLLM used in \methodNameNS~is GPT-4o~\cite{achiam2023gpt}, while we also perform ablations with GPT-4o-mini, GPT-4V, and open-source MLLMs including LLaVA~\cite{liu2024visual} and MiniGPT4 \cite{zhu2023minigpt}. GPT APIs are used with a temperature setting of 0, while all other parameters remain at their default values. The retrieval module, built in PyTorch~\cite{paszke2019pytorch} based on the codebase from \cite{training-free-cir}, performs all computations on a single NVIDIA A100 GPU. For the CLIP-based ViT variants~\cite{dosovitskiy2020image}, we adopt weights from the official CLIP implementation~\cite{radford2021learning} while using OpenCLIP~\cite{openclip} for ViT-G/14. Performance metrics are averaged across three trials to ensure reliability.

\subsection{Quantitative and Qualitative Results}

\begin{figure}[t]
    \centering
    \centering
    \includegraphics[width=1.0\linewidth]{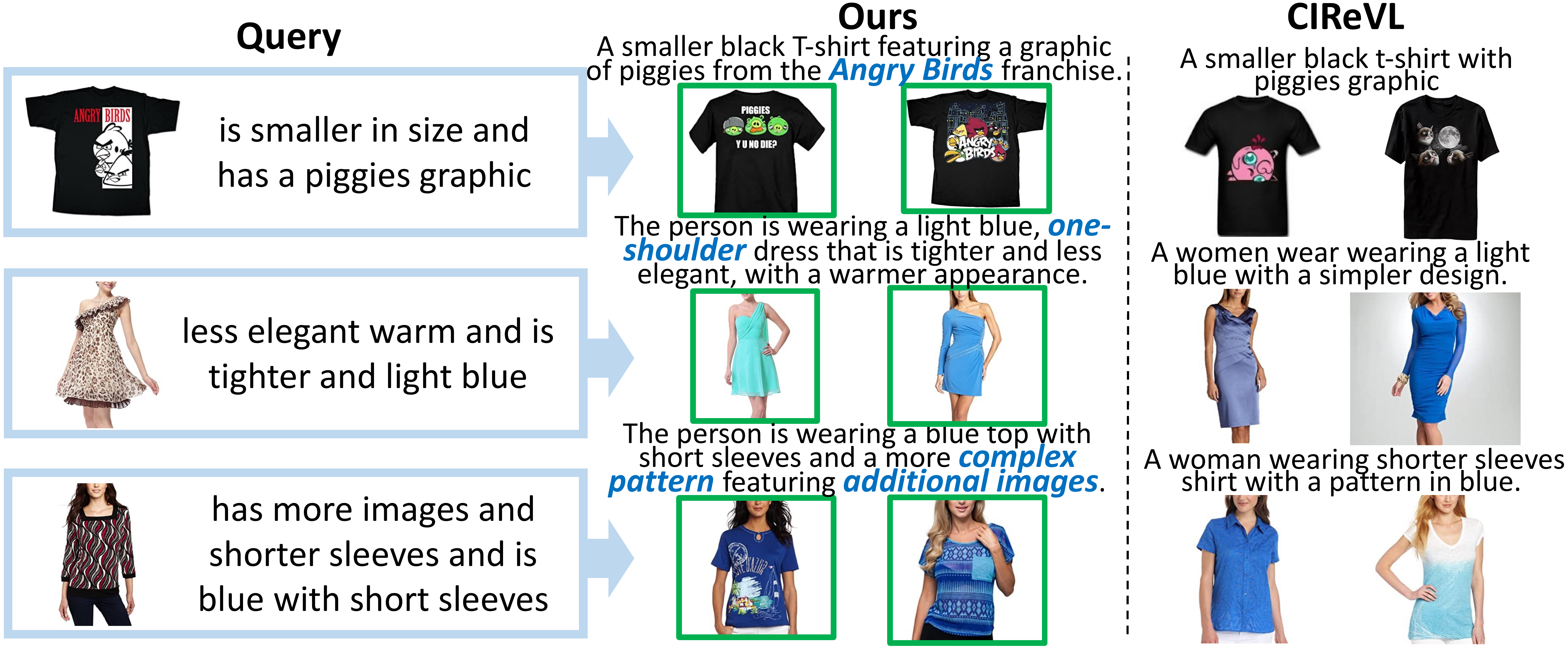}
    \caption{Results of attribute manipulation on the FashionIQ.}
   \vspace{-5pt}
    \label{fig:fashion}
    \vspace{-10pt}
\end{figure}

Our main quantitative experimental results are presented in  Tables~\ref{tab:circo}, \ref{tab:genecis} and \ref{tab:fiq}, while Figures \ref{fig:cirr} and \ref{fig:fashion} show qualitative comparisons between our model and the baseline CIReVL.

In Table \ref{tab:circo}, we show the comparison results for the CIRCO and CIRR datasets, which evaluate our model's capability in foreground and background differentiation as well as fine-grained image editing through object and scene manipulation tasks. Performances are evaluated on the hidden test sets of CIRCO and CIRR, accessible via the submission servers~\cite{baldrati2023zero, Saito_2023_CVPR}. For all different CLIP-based ViT variants for retrieval, our approach significantly outperforms existing methods, including training-free and textual inversion. For instance, on the default ViT-L/14 in CIRCO, which contains clean annotations of manipulation text with multiple target images, our model achieves a mAP@5 of 23.87\%, notably surpassing the 18.92\% obtained by the best training-free method (CIReVL$^*$) and nearly doubling the 13.04\% achieved by the top textual inversion method (Context-I2W). 
Furthermore, in CIRR, where the manipulation text is less explicit and noisier \cite{baldrati2023zero, karthik2024visionbylanguage}, our model still shows a significant 3.23\% average improvement across all evaluation metrics over the best training-free method, CIReVL$^*$. Note that although CIReVL$^{*}$ outperforms CIReVL, the difference is marginal, suggesting that simply adopting a better MLLM does not address the limitations of the two-stage approach.

Qualitatively, as illustrated in Figure~\ref{fig:cirr}, our method, \methodNameNS, generates target image descriptions that align with user intent and capture intricate visual details. In comparison, CIReVL misses critical elements, such as the image type ``poster" and dog breed ``Chihuahuas" in Row 1, the dog's ``tan" color in Row 2, and the contextual details of the ``beach" setting and dog breed ``Labrador" in Row 3.

\begin{table*}[t!]
    \centering
    \resizebox{0.7\linewidth}{!}{
    \begin{tabular}{l|l|cccccc|cc}
    \toprule
    \multicolumn{2}{c|}{\textbf{Fashion-IQ} $\rightarrow$} & \multicolumn{2}{c}{Shirt} & \multicolumn{2}{c}{Dress} & \multicolumn{2}{c}{Toptee} & \multicolumn{2}{|c}{\textbf{Average}}\\
    \midrule
     Backbone & Method & R@10 & R@50 & R@10 & R@50 & R@10 & R@50 & R@10 & R@50 \\
     \midrule
     \multirow{4}{*}{ViT-B/32} & SEARLE & 24.44 & 41.61 & 18.54 & 39.51 & 25.70 & 46.46 & 22.89 & 42.53 \\
       & \greycell CIReVL & \greycell 28.36 & \greycell 47.84 & \greycell 25.29 & \greycell 46.36 & \greycell 31.21 & \greycell 53.85 & \greycell 28.29 & \greycell 49.35 \\
      & \greycell CIReVL$^{*}$ & \greycell \underline{28.83} & \greycell \underline{48.36} & \greycell \underline{25.82} & \greycell \underline{46.89} & \greycell \underline{31.73} & \greycell \underline{54.34} & \greycell \underline{28.79} & \greycell \underline{49.86} \\
       & \greycell \textbf{\methodNameNS} & \greycell \textbf{31.16} & \greycell \textbf{51.13} & \greycell \textbf{29.35} & \greycell \textbf{50.37} & \greycell \textbf{36.51} & \greycell \textbf{58.71} & \greycell \textbf{32.34} & \greycell \textbf{53.40} \\
    \midrule
     \multirow{7}{*}{ViT-L/14}& Pic2Word & 26.20 & 43.60 & 20.00 & 40.20 & 27.90 & 47.40 & 24.70 & 43.70 \\
      & SEARLE & 26.89 & 45.58 & 20.48 & 43.13 & 29.32 & 49.97 & 25.56 & 46.23\\
      & LinCIR & 29.10 & 46.81 & 20.92 & 42.44 & 28.81 & 50.18 & 26.28 & 46.49\\
      & Context-I2W & 29.70 & \underline{48.60} & 23.10 & \underline{45.30} & 30.60 & 52.90 & 27.80  & 48.90\\
      & \greycell CIReVL & \greycell 29.49 & \greycell 47.40 & \greycell 24.79 & \greycell 44.76 & \greycell 31.36 & \greycell 53.65 & \greycell 28.55 & \greycell 48.57\\ 
      & \greycell CIReVL$^{*}$ & \greycell \underline{29.98} & \greycell  47.92 & \greycell \underline{25.29} & \greycell 45.28 & \greycell \underline{31.89} & \greycell \underline{54.13} & \greycell \underline{29.05} & \greycell \underline{49.11}\\ 
      & \greycell \textbf{\methodNameNS} & \greycell \textbf{33.17} & \greycell \textbf{52.03} & \greycell \textbf{29.70} & \greycell \textbf{51.81} & \greycell \textbf{36.92} & \greycell \textbf{59.27} & \greycell \textbf{33.26} & \greycell \textbf{54.37} \\
     \midrule
     \multirow{4}{*}{ViT-G/14}& LinCIR & \textbf{46.76} & \textbf{65.11} & \textbf{38.08} & \textbf{60.88} & \textbf{50.48} & \textbf{71.09} & \textbf{45.11} & \textbf{65.69} \\
      & \greycell CIReVL & \greycell 33.71 & \greycell 51.42 & \greycell 27.07 & \greycell 49.53 & \greycell 35.80 & \greycell 56.14 & \greycell 32.19 & \greycell 52.36\\
    & \greycell CIReVL$^{*}$ & \greycell 34.01 & \greycell 51.92 & \greycell 27.56 & \greycell 50.04 & \greycell 36.29 & \greycell 56.63 & \greycell 32.62 & \greycell 52.86\\
     & \greycell \textbf{\methodNameNS} & \greycell \underline{38.65} & \greycell \underline{54.71} & \greycell \underline{33.02} & \greycell \underline{54.78} & \greycell \underline{41.04} & \greycell \underline{61.83} & \greycell \underline{37.57} & \greycell \underline{57.11} \\ 
    \bottomrule
    \end{tabular}}    
    \vspace{-5pt}
    \caption{\textbf{Comparison on FashionIQ Validation Data.} \methodNameNS~is able to significantly outperform adaptive methods across all sub-benchmarks, with its inherent modularity allowing for further simple scaling to achieve additional large gains. Grey lines represent the training-free ZS-CIR methods. CIReVL$^*$ uses the GPT4o in two-stage. \textbf{Bold} and `$\underline{\phantom{x}}$' denotes the best and second-best result, respectively.}
    \label{tab:fiq}
    \vspace{-12pt}
\end{table*}

We further evaluate our model's capability on object and attribute composition using the GeneCIS dataset, with the results detailed in Table~\ref{tab:genecis}. Unlike CIRCO and CIRR, GeneCIS uses single-word manipulation texts with varied interpretations depending on the task, such as focusing on or changing a specific attribute or object. Consequently, user intent is often abstract and ambiguous, requiring our model to interpret intent precisely based on the reference image. For a fair comparison, we adopt the same output format in our reflective CoT process as the recent work~\cite{karthik2024visionbylanguage}. Specifically, for the ``Focus'' tasks, we direct the MLLM to \textit{retain the attribute or object specified in the instruction}. For the ``Change'' tasks, we prompt it to \textit{replace the corresponding object}. For the ViT-L/14 retrieval backbone, our method achieves a 1.8\% improvement in Average $R@1$ over the best training-free method (CIReVL$^*$) and outperforms the best textual inversion method (Context-I2W) by 5.2\%. Similar improvements are also observed for the other two backbones, underscoring the effectiveness of our reflective CoT process in capturing the user's implicit intent.

Lastly, Table \ref{tab:fiq} presents our model’s performance on attribute manipulation tasks using the FashionIQ validation set, requiring accurate localization of specific fashion attributes (\textit{e.g.,} style, color, pattern). The results show that \methodNameNS~surpasses existing ZS-CIR models using the ViT-B/14 and ViT-L/14 backbones. For instance, on ViT-L/14, our method outperforms the best training-free model (CIReVL$^*$) and the leading textual inversion model (Context-I2W) by 4.74\% and 5.47\% on average, respectively. 
On ViT-G/14, our method achieves a notable 4.6\% improvement over the best training-free model, CIReVL$^*$, yet still falls short of the best-performing textual inversion approach, LinCIR. This discrepancy may stem from LinCIR’s training process being aligned with the CLIP model used in retrieval, unlike our training-free approach, which lacks this specific alignment. The limitation is particularly evident in the fashion domain, where CLIP may have limited domain-specific knowledge. For instance, terms like ``sequined bodice” in the target description are challenging for CLIP to interpret without training-based alignment, leading to reduced performance. Conversely, in the natural image domain, such as CIRCO, where MLLM/LLM outputs are more comprehensible to CLIP, our training-free method substantially outperforms all textual inversion techniques. Future work might explore enhancing the alignment between reasoning and retrieval modules to improve model performance in specialized domains.

Qualitative comparison results of our method and the baseline method CIReVL are presented in Figure~\ref{fig:fashion}. \methodNameNS~accurately identifies and manipulates the attribute-relevant visual elements of ``Angry Birds'' (Row 1), a ``one-shoulder'' dress (Row 2), and a tee with a complex pattern featuring more images (Row 3). 

\subsection{Ablation Study and Performance Analysis}
\label{sec::anaysis}

\begin{figure*}[t]
    \centering
    \includegraphics[width=1.0\linewidth]{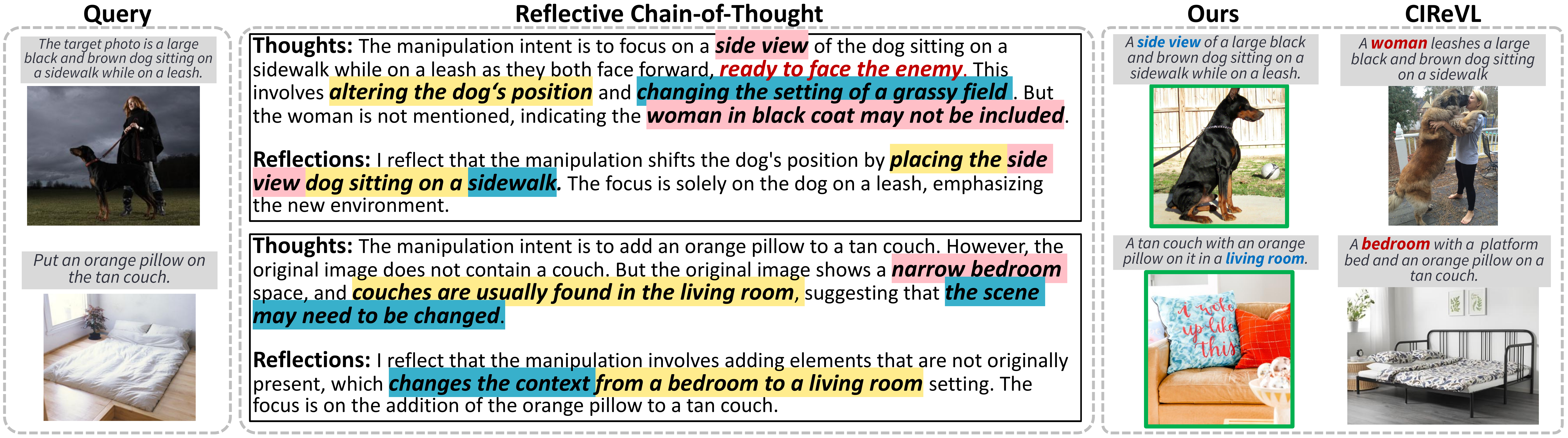}
    \vspace{-15pt}
    \caption{Visualization of Reflective CoT samples. We compare the top 1 retrieval results of ours and CIReVL. Different colors denote the reasoning outcomes of each intention. Our Reflective CoT effectively filters out elements irrelevant to user intention.}
    \label{fig:CoT}
    \vspace{-15pt}
\end{figure*}

Similar to \cite{karthik2024visionbylanguage,yang2024ldre, gu2024lincir}, we examine the contributions of core components in \methodNameNS~using a ViT-L/14 backbone on CIRCO and Fashion-IQ (Table \ref{tab:ablation}). \textbf{(1) Models `2-3' assess the impact of key modules in \methodNameNS.} Adapting CIReVL's second stage with our reflective CoT process (model `2') results in a 2.46\% average performance drop compared to our method (model `1'), highlighting the necessity of our one-stage reasoning process for capturing complete reference image content. Removing Reflective CoT (model `3') causes a 3.55\% performance decline, indicating the importance of our multimodal CoT for effective manipulation intention understanding. We choose not to conduct an ablation integrating manipulation text into caption generation with MLLM in the two-stage approach, as it is methodologically closely aligned with \methodNameNS~but adds an additional MLLM query, which is unnecessary and reduces efficiency. \textbf{(2) Models `4-7' evaluate each Reflective CoT step.} Skipping the generation of the original image description guided by manipulation text (model `4') causes a 1.44\% performance decline, emphasizing the need to remove irrelevant visual information. Similarly, without reasoning about user intentions (model `5') or filtering irrelevant ones (model `6'), performance drops by 2.60\% and 2.08\%, respectively, underscoring the importance of capturing user intentions and identifying relevant visual elements. Removing our vision-by-language in-context samples (model `7') results in a 1.29\% decline, showing the benefit of ICL for guiding the reflective CoT. \textbf{(3) In models `8-11', we analyze the impact of the choice of MLLM.} Open-source models, such as LLaVA (model `8') and MiniGPT-4 (model `9'), achieve results close to the best training-free ZS-CIR method, CIReVL,  but there remains a gap of 2.89\% and 3.96\% compared to GPT-4o (model `1'). Notably, GPT-4o-mini (model `10') performs comparably well, with only a 0.97\% decline while being more efficient than GPT-4o.

\begin{table}[t]

\centering
\scalebox{1.0}
{
\footnotesize
\setlength{\tabcolsep}{1.0mm}
\begin{tabular}{llcclcc}
\toprule
\multicolumn{2}{l}{}   & \multicolumn{3}{c}{CIRCO}                         & \multicolumn{2}{c}{Fashion-IQ} \\ \cmidrule(lr){3-5}\cmidrule(lr){6-7}
      & Methods           & k=5   & k=10   & k=25                                & k=10            & k=50           \\ \cmidrule(lr){1-7}
1. & Full model (GPT-4o)        & 23.87 & 25.33 & \multicolumn{1}{l|}{27.84}          & 33.26           & 54.37           \\
\multicolumn{5}{l}{\textbf{Significance of key modules of \methodNameNS}} & & \\ 
2. & w/o one-stage reasoning       & 21.73 & 22.78 & \multicolumn{1}{l|}{24.47}          & 31.16           & 52.22         \\
3. & w/o Reflective CoT       & 20.86 & 21.40 & \multicolumn{1}{l|}{23.34}          & 30.27           & 51.06         \\
\multicolumn{5}{l}{\textbf{Necessity of each step in our Reflective CoT}} & & \\ 
4. & w/o Original Description       & 22.56 & 23.57 & \multicolumn{1}{l|}{26.02}          & 32.37           & 52.97         \\
5. & w/o Thoughts       & 21.46 & 22.07 & \multicolumn{1}{l|}{25.06}          & 31.59           & 51.47         \\
6. & w/o Reflections         & 22.04 & 22.74 & \multicolumn{1}{l|}{25.32}        & 32.05           & 52.11         \\
7. & w/o ICL       & 22.97 & 23.50 & \multicolumn{1}{l|}{26.55}          & 32.03           & 53.17        \\
\multicolumn{5}{l}{\textbf{Impact of different MLLMs}} & & \\ 
8. & LLaVA       & 20.89 & 22.30 & \multicolumn{1}{l|}{24.88}          & 30.75           & 51.42         \\
9. & MiniGPT-4       & 19.85 & 21.30 & \multicolumn{1}{l|}{23.90}          & 29.36           & 50.47         \\
10. & GPT-4o-mini       & 23.10 & 24.47 & \multicolumn{1}{l|}{26.73}          & 32.19           & 53.32         \\
11. & GPT-4V       & 22.15 & 23.58 & \multicolumn{1}{l|}{25.24}          & 31.55           & 52.60         \\

\bottomrule
\end{tabular}}
\caption{Ablation study on CIRCO and FashionIQ.}
\label{tab:ablation}
\vspace{-15pt}
\end{table}

\paragraph{Qualitative Analysis of Reflective CoT.} To further examine the benefits of reflective CoT on interpreting user intent, we present additional case studies in Figure~\ref{fig:CoT} alongside the example in Figure~\ref{fig:model-architecture}. For instance, in Row 1, reflective CoT effectively filters out elements irrelevant to user intent, such as ``the woman in a black coat'' and the hallucinated thought (\textit{i.e.,} ``ready to face the enemy''). Notably, reflective CoT also demonstrates accuracy in interpreting intent even when the connection between the reference image and manipulation text is weak, as shown in Row 2. Although this situation technically falls outside CIR, it reflects real user behavior, where users may not closely align manipulation text with the reference image. In Row 2, reflective CoT uses common sense (\textit{e.g.}, recognizing that couches are uncommon in small bedrooms) to infer the user’s intention of transitioning from a bedroom to a living room. This filtering of irrelevant details enhances model robustness and likely underlies its strong performance on the CIR task. 

\begin{figure}[t]
    \centering
    \includegraphics[width=1.0\linewidth]{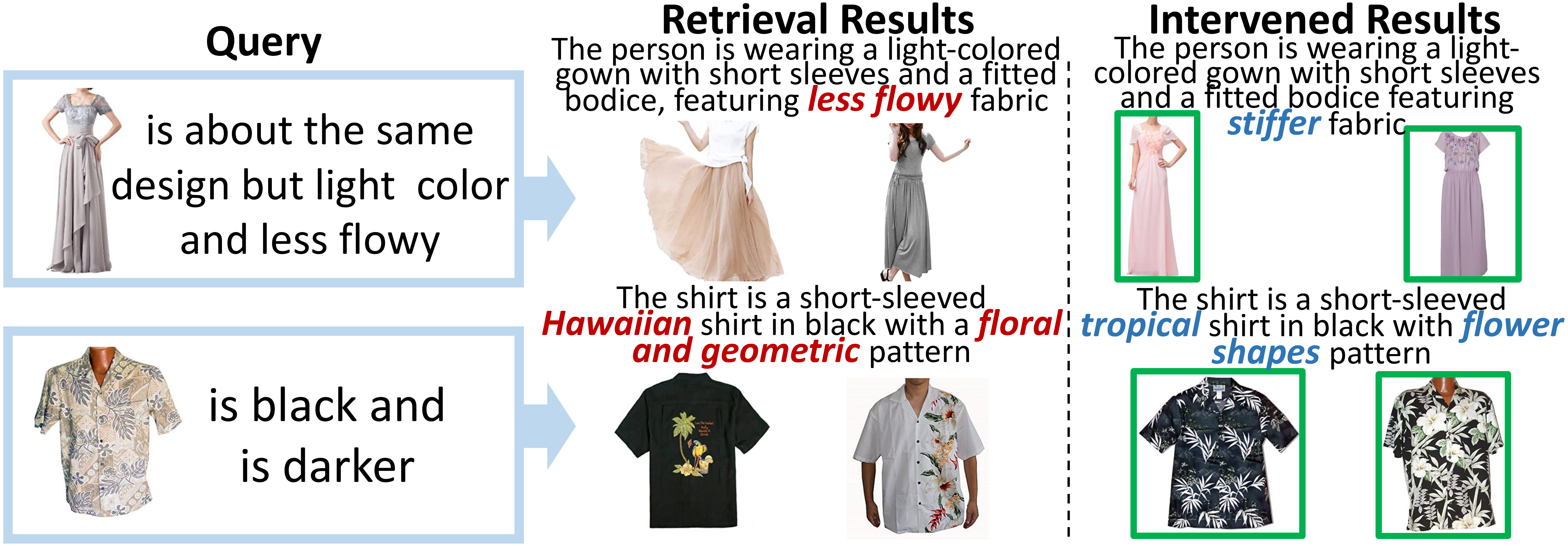}
    \vspace{-15pt}
    \caption{Visualization of common failure cases in the FashionIQ validation set. The top 2 retrieval results are shown.}
    \label{fig:fail}
    \vspace{-15pt}
\end{figure}

\noindent\textbf{Analysis of Common Failure Cases.} To gain insights into failure cases of \methodNameNS, we analyzed 300 failure cases from the FashionIQ validation set using ViT-G/14. As shown in Figure~\ref{fig:fail}, we identify two main issues: (1) \textit{Difficulty with reasoning terms} (49\%): The retrieval model (\textit{i.e.,} CLIP) often misreads reasoning terms (\textit{e.g.,} comparisons) like interpreting ``less flowy'' incorrectly (Row 1) while substituting ``stiffer'' corrected this. (2) \textit{Misalignment concepts between MLLM and retrieval model} (34\%): the retrieval model struggles to interpret fashion-specific terms from MLLM, like ``Hawaiian style'' and ``floral and geometric'' (Row 2). Replacing them with simpler terms (``tropical style'', ``flower shapes'') improved retrieval accuracy. 

\noindent\textbf{Effectiveness and Efficiency Analysis.} Our approach not only outperforms the best training-free ZS-CIR method (CIReVL) on four ZR-CIR tasks, and also has faster inference time, taking about $0.6$ second per query that is 66.67\% faster than CIReVL. Compared to textual inversion methods, while our performance surpasses them without training, our inference speed remains \(30\times\) slower. As MLLM API calls account for 97\% of the total time in \methodNameNS, we believe that faster APIs may resolve this issue in the future. 

\vspace{-20pt}
\section{Conclusion}
\vspace{-5pt}
In this paper, we propose a one-stage reflective chain-of-thought reasoning approach that leverages MLLMs to simultaneously process visual and textual inputs, reducing information loss found in two-stage training-free ZS-CIR methods. By capturing nuanced manipulation intents from text and image cues, \methodNameNS~demonstrates strong generalization and significantly outperforms existing methods on four diverse tasks, achieving comparable inference times. This work advances intention-based image retrieval and has broad implications for vision-language applications.

{
    \small
    \bibliographystyle{ieeenat_fullname}
    \bibliography{main}
}
\clearpage
\renewcommand{\thesection}{\Alph{section}}
\setcounter{section}{0}

\section{Complete Template for Reflective CoT}
\label{sec::tem}
\begin{figure*}[h]
    \centering
    \includegraphics[width=1.0\linewidth]{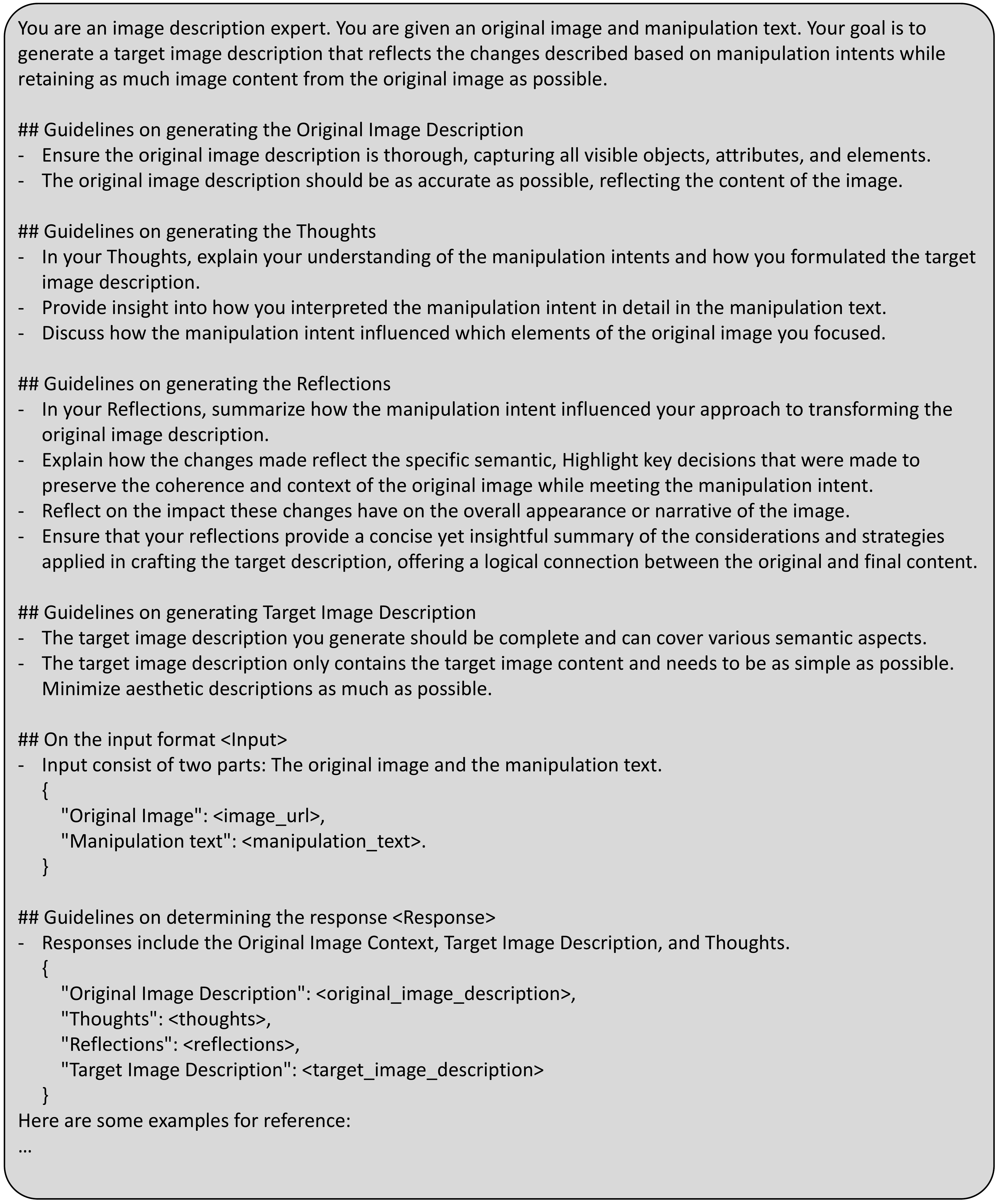}
    \caption{The complete template of our reflective Chain-of-Thought process for Training-free ZS-CIR.}
    \label{fig:vis_prompt}
\end{figure*}

The complete template of our reflective CoT prompt is shown in Figure~\ref{fig:vis_prompt}. The Reflective CoT prompt instructs the following progressive reasoning steps: First, the \textit{Original Image Description} step highlights visual details relevant to the user's intention in the reference image. The \textit{Thoughts} step then captures the user's intention and reasoning for potentially manipulated visual elements. In the \textit{Reflections} step, these elements are further evaluated to identify those mostly aligned with the user's intent. Finally, the \textit{Target Image Description} step generates a refined description based on the most intention-relevant visual modifications for target retrieval. Notably, all steps are included in a \textbf{single} prompt for MLLM, ensuring both efficiency and interpretability.  

\vspace{10pt}

\noindent\textbf{Original Image Description.} ~During this step, the MLLM is asked to \textit{capture all visible objects, attributes, and elements relevant to the manipulation text}, and to \textit{reflect on the content and context of the image} to ensure retention of fine-grained details. 

\vspace{10pt}

\noindent\textbf{Thoughts.} Given the intention-relevant visual details and manipulation text, the MLLM then seeks to capture the user's intention. We first prompt the MLLM to \textit{explain its understanding of the manipulation intent}. Since the user’s intentions are often implicit, requiring reference image context for interpretation, we further ask the MLLM to \textit{discuss how the manipulation intent influences the choice of focused elements in the original image}.

\vspace{10pt}

\noindent\textbf{Reflections.} Given the manipulation intent and reference image, the MLLM needs to filter out incorrect intentions and identify the most relevant manipulated elements. We ask the MLLM to \textit{highlight key decisions made to preserve the coherence and context of the original image while fulfilling the manipulation intent} and to \textit{offer a logical connection between the original content and the final description.} 

\vspace{10pt}

\noindent\textbf{Target Image Description.} Given the manipulated visual elements most relevant to the user's intention, the AI agent needs to generate a target description that associates those manipulated visual elements for retrieval. We simply ask the MLLM to \textit{generate a target image description that only contains the target image content}.

\paragraph{Input and Output.} As shown in Figure \ref{fig:vis_prompt}, the input to the LLM is a concatenated prompt as \( T_{t} = \Psi_M(p_c \circ I_r \circ T_{m})\)  comprising the base CoT prompt \( p_{c} \), the base64-encoded image URL of the reference image \( I_r \) (prepended with ``\texttt{Original Image Context}''), and the manipulation intent text \( T_{m} \) (prepended with ``\texttt{Manipulation Text}''). This task-agnostic prompt format allows for application across various CIR tasks. The output is provided as a JSON file containing ``Original Image Description'', ``Thoughts'', ``Reflections'', and ``Target Image Description''. The ``Target Image Description'' is selected as the final output, while the additional information can serve as valuable reference data for LLM-based ensemble methods \cite{yang2024ldre}, potentially boosting performance at the cost of efficiency.

\begin{figure*}[h]
    \centering
    \includegraphics[width=1.0\linewidth]{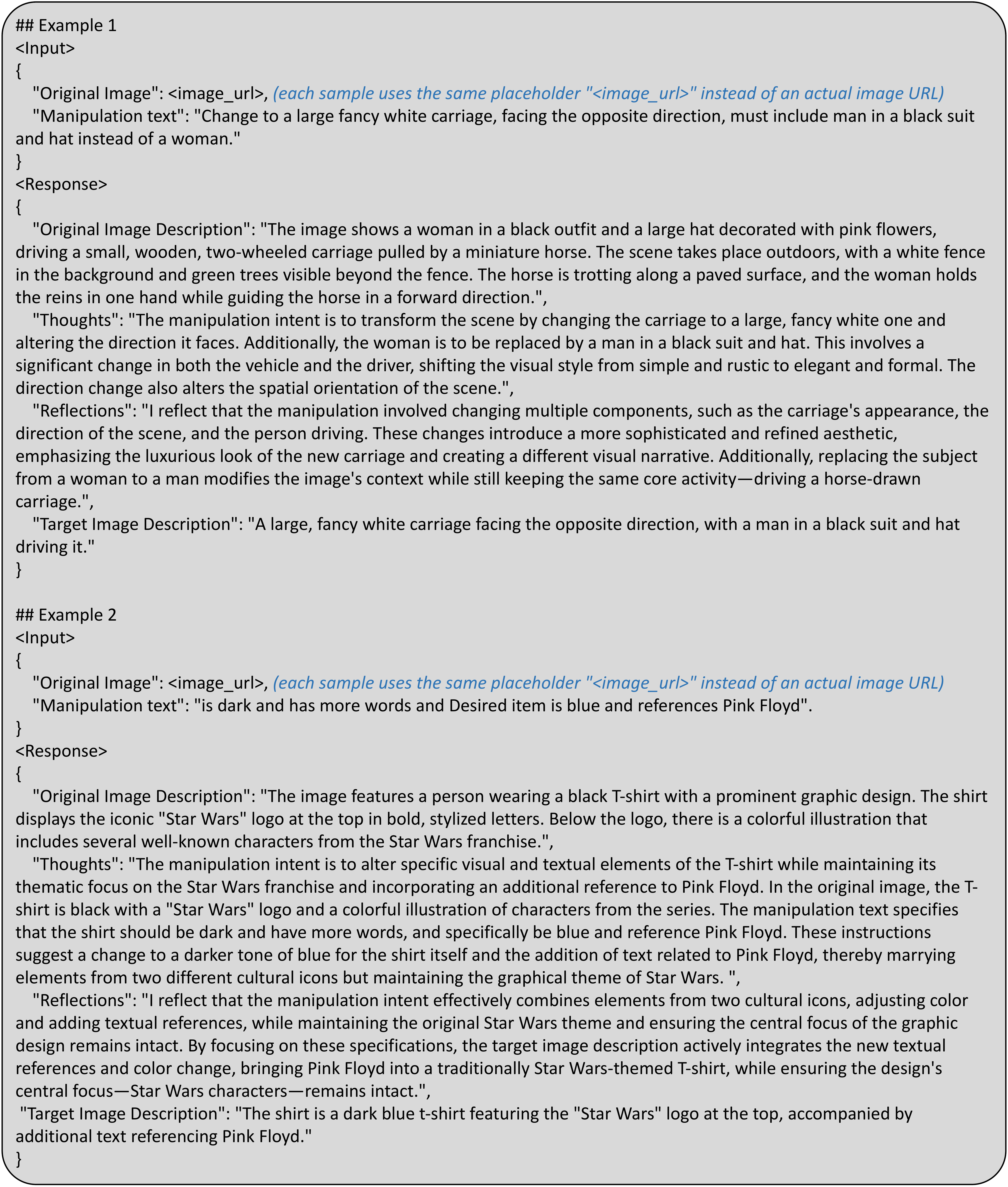}
    \caption{Examples of our vision-by-language in-context learning. Notably, each sample uses the same placeholder ``\texttt{<image\_url>}'' instead of an actual reference image URL, guiding the MLLM formatting of the input and output. }
    \label{fig:vis_ICTt}
\end{figure*}

\section{Vision-by-Language In-Context Learning Details} 
\label{sec::ICT_details}
Simply providing guidelines for the Reflective CoT process is insufficient for MLLMs to understand the CoT process required at each step. To address this, we leverage in-context learning, a technique widely used in LLM and MLLM CoT methods \cite{wei2022chain, mitra2023compositional, zheng2023ddcot}. 


To ensure a zero-shot setting in ZS-CIR, we propose a vision-by-language in-context learning (ICL) approach. As illustrated in Figure \ref{fig:vis_ICTt}, our vision-by-language ICL provides a few expected MLLM outputs (\textit{i.e.,} three samples) in text form as examples, without requiring a reference image to guide the MLLM through the reasoning process at each step. Notably, each sample uses the same placeholder ``\texttt{<image\_url>}'' instead of an actual reference image URL, guiding the MLLM formatting of the input and output. 

For instance, consider the manipulation text (sample 1): \textit{``Change to a large fancy white carriage, facing the opposite direction, must include man in a black suit and hat instead of a woman.''} The language-based description of the reference image is: \textit{``The image shows a woman in a black outfit and a large hat decorated with pink flowers, driving a small, wooden, two-wheeled carriage pulled by a miniature horse.''} Following the Reflective CoT steps:

\begin{itemize}
    \item \textbf{Original Image Description:} The MLLM captures all visible objects and attributes relevant to the manipulation text, ensuring fine-grained details are included. In this case, it notes the woman in a black outfit with a large hat, the small wooden carriage, the miniature horse, and the outdoor setting with a white fence and trees.
    
    \item \textbf{Thoughts:} The MLLM interprets the manipulation intent by explaining that the scene should be transformed into one featuring a large, fancy white carriage facing the opposite direction, and the woman replaced with a man in a black suit and hat. This step demonstrates the model's understanding of the required changes and how they influence the focused elements in the original image.
    
    \item \textbf{Reflections:} The MLLM reflects on key decisions to preserve coherence while fulfilling the manipulation intent. It acknowledges that changing multiple components—such as the carriage's appearance, the direction it faces, and the driver—introduces a more sophisticated aesthetic and alters the visual narrative. The reflection offers a logical connection between the original content and the final description.
    
    \item \textbf{Target Image Description:} The MLLM generates a refined description containing only the target image content: \textit{``A large, fancy white carriage facing the opposite direction, with a man in a black suit and hat driving it.''}
\end{itemize}

This example illustrates how our vision-by-language in-context learning approach guides the MLLM through each step of the Reflective CoT process, enabling it to produce accurate and coherent descriptions for the target image without direct visual input. By providing language-based examples, the MLLM can internalize the reasoning pattern and apply it to new instances, ensuring consistency and effectiveness in zero-shot settings without reference images.

\section{More Ablation Study} 
\label{sec::more_abl}

Table \ref{tab:more_ablation} presents additional ablation analyses. \textbf{(1) Models `2-4' assess the significance of the one-stage reasoning strategy.} Using GPT-4o as the captioner with manipulation text to enhance the reference image captioning process (model `2') results in a 3.62\% performance decline, while incorporating GPT-4o with our Reflective CoT process (model `3') leads to a 2.46\% decline. These results highlight the necessity of our one-stage reasoning process for capturing complete reference image content and the importance of multimodal CoT for effective manipulation intention understanding. Incorporating manipulation text into caption generation in the two-stage approach (model `4') achieves similar performance but introduces additional MLLM queries, reducing efficiency, and is therefore unnecessary. \textbf{(2) Models `5-6' evaluate different backbone retrieval models.} \methodNameNS~with BLIP ViT-L/16 \cite{pmlr-v162-li22n} and Long-CLIP ViT-L/14 \cite{zhang2024longclip} achieves results comparable to the CLIP, demonstrating the generalizability and robustness of \methodNameNS~across different CLIP-based backbones.

\begin{table}[t]

\centering
\scalebox{0.96}
{
\footnotesize
\setlength{\tabcolsep}{0.8mm}
\begin{tabular}{llcclcc}
\toprule
\multicolumn{2}{l}{}   & \multicolumn{3}{c}{CIRCO}                         & \multicolumn{2}{c}{Fashion-IQ} \\ \cmidrule(lr){3-5}\cmidrule(lr){6-7}
      & Methods           & k=5   & k=10   & k=25                                & k=10            & k=50           \\ \cmidrule(lr){1-7}
1. & Full model (GPT-4o)        & 23.87 & 25.33 & \multicolumn{1}{l|}{27.84}          & 33.26           & 54.37           \\
\multicolumn{5}{l}{\textbf{Significance of the one stage reasoning strategy}} & & \\ 
2. & two-stage+enhance captioner        & 20.93 & 21.34 & \multicolumn{1}{l|}{23.27}          & 30.14           & 50.87         \\
3. & two-stage+CoT       & 21.73 & 22.78 & \multicolumn{1}{l|}{24.47}          & 31.16           & 52.22         \\
4. & two-stage+enhance captioner+CoT        & 23.24 & 24.97 & \multicolumn{1}{l|}{27.04}          & 32.54           & 53.47         \\
\multicolumn{5}{l}{\textbf{Impact of different backbone models}} & & \\ 
5. & BLIP       & 23.93 & 25.47 & \multicolumn{1}{l|}{27.53}          & 32.10           & 53.69        \\
6. & long clip        & 23.73 & 25.12 & \multicolumn{1}{l|}{26.91}          & 31.77           & 53.02         \\


\bottomrule
\end{tabular}}
\caption{More Ablation study on CIRCO and FashionIQ.}
\label{tab:more_ablation}
\vspace{-15pt}
\end{table}

\begin{algorithm}[tb]
\caption{Computing Process of \methodNameNS}
\label{alg:algorithm_reflective_cot}
\textbf{Input}: Reference image \( I_r \), manipulation text \( T_m \), reflective CoT prompt \( p_c \), image-search database \( \mathcal{D} \).\\
\textbf{Parameters}: Frozen MLLM \( \Psi_M \), frozen CLIP vision encoder \( \Psi_I \), language encoder \( \Psi_T \).\\
\textbf{Output}: Retrieved target image \( I_t \).

\begin{algorithmic}[1]
\STATE Initialize pre-trained and frozen models \( \Psi_M \), \( \Psi_I \), \( \Psi_T \).
\STATE Generate target image description: \\
\[ T_t = \Psi_M(p_c \circ I_r \circ T_m) \]
\STATE Compute normalized text embedding: 
\[ \hat{e}_T = \frac{\Psi_T(T_t)}{\|\Psi_T(T_t)\|} \]
\FOR{each image \( I_i \) in \( \mathcal{D} \)}
    \STATE Compute normalized image embedding: \\
    \[ \hat{e}_{I_i} = \frac{\Psi_I(I_i)}{\|\Psi_I(I_i)\|} \]
    \STATE Compute similarity score: \( s_i = \hat{e}_{I_i}^\top \hat{e}_T \)
\ENDFOR
\STATE Retrieve target image: \( I_t = \underset{I_i \in \mathcal{D}}{\mathrm{argmax}} \ s_i \)
\STATE \textbf{return} \( I_t \)
\end{algorithmic}
\label{alg:algorithm}
\end{algorithm}

\section{Algorithm of \methodNameNS's Process.} 

Algorithm \ref{alg:algorithm} outlines \methodNameNS's process for training-free ZS-CIR. Given the target image description \( T_t \), the model encodes the image-search database \( \mathcal{D} \) and \( T_t \) using a frozen pre-trained CLIP. The retrieved target image \( I_t \) is selected based on cosine similarity \( \texttt{cos}(\Psi_I(I_c), \Psi_T(T_t)) \), where \( I_t \) is the image most similar to the generated description \( T_t \). This retrieval process is modular and performed after combining the reference image and manipulation text, allowing for flexible substitution of retrieval systems to balance efficiency and effectiveness. The approach creates a human-understandable ZS-CIR pipeline, fully expressing reasoning in the language domain while keeping the retrieval process independent, requiring no additional training modules.

\end{document}